%% file: 01722.tex
\RequirePackage{amsmath}
\documentclass[runningheads]{llncs}

\usepackage[T1]{fontenc}

\usepackage{graphicx}
\usepackage[dvipsnames]{xcolor}
\usepackage{amsmath,graphicx}

\usepackage{soul}

\renewcommand{\hl}[1]{#1}

\usepackage{array}
\usepackage{booktabs}
\usepackage{makecell}
\usepackage{siunitx}
\usepackage{multirow}

\usepackage{siunitx}
\usepackage[misc]{ifsym}

\def\x{$\times$}

\begin{document}

\title{Detection of Intracranial Hemorrhage \\for Trauma Patients}

\author{Antoine P. Sanner\inst{1,2}\orcidID{0000-0002-4917-9529} \textsuperscript{(\Letter)} \and
Nils F. Grauhan\inst{2} \and
Merle Meyer\inst{2} \and
Laura Leukert\inst{2} \and
Marc A. Brockmann\inst{2} \and
Ahmed E. Othman\inst{2} \and
Anirban Mukhopadhyay\inst{1}\orcidID{0000-0003-0669-4018}}

\index{Antoine Sanner}
\index{Nils Grauhan}
\index{Merle Meyer}
\index{Laura Leukert}
\index{Marc Brockmann}
\index{Ahmed Othman}
\index{Anirban Mukhopadhyay}

\authorrunning{A. Sanner et al.}

\institute{Department of Computer Science, Technical University of Darmstadt, Germany \\ 
\and
Department of Neuroradiology, University Medical Center Mainz, Germany \\
\email{antoine.sanner@gris.tu-darmstadt.de}
}

\maketitle              

\input{Chapters/0_abstract}

\input{Chapters/1_intro}
\input{Chapters/2_methods}
\input{Chapters/3_data}
\input{Chapters/4_exp}
\input{Chapters/5_results}
\input{Chapters/6_discussion}
\input{Chapters/7_ethics}

\begin{credits}
\subsubsection{\discintname}
The authors have no competing interests to declare that are relevant to the content of this article.
\end{credits}

\bibliographystyle{splncs04}
\bibliography{01722}
\end{document}

%% file: Chapters/0_abstract.tex
\begin{abstract}

Whole-body CT is used for multi-trauma patients in the search of any and all injuries. 
Since an initial assessment needs to be rapid and the search for lesions is done for the whole body, very little time can be allocated for the inspection of a specific anatomy.
In particular, intracranial hemorrhages are still missed, especially by clinical students. In this work, we present a Deep Learning approach for highlighting such lesions to improve the diagnostic accuracy.
While most works on intracranial hemorrhages perform segmentation, detection only requires bounding boxes for the localization of the bleeding. 
In this paper, we propose a novel Voxel-Complete IoU (VC-IoU) loss that encourages the network to learn the 3D aspect ratios of bounding boxes and leads to more precise detections. We extensively experiment on brain bleeding detection using a publicly available dataset, and validate it on a private cohort, where we achieve 0.877 AR$_{30}$, 0.728 AP$_{30}$ and 0.653 AR$_{30}$, 0.514 AP$_{30}$ respectively. These results constitute a relative +5\% improvement in Average Recall for both datasets compared to other loss functions.
Finally, as there is little data currently publicly available for 3D object detection and as annotation resources are limited in the clinical setting, we evaluate the cost of different annotation methods, as well as the impact of imprecise bounding boxes in the training data on the detection performance. 

\end{abstract}

\begin{keywords}
3D voxel object detection, Intracranial Hemorrhage, Multi-Trauma
\end{keywords}

%% file: Chapters/1_intro.tex
\section{Introduction}
\label{sec:intro}

\begin{figure}[htb]

\begin{minipage}[b]{1.0\linewidth}
  \centering
  \centerline{\includegraphics[width=11cm]{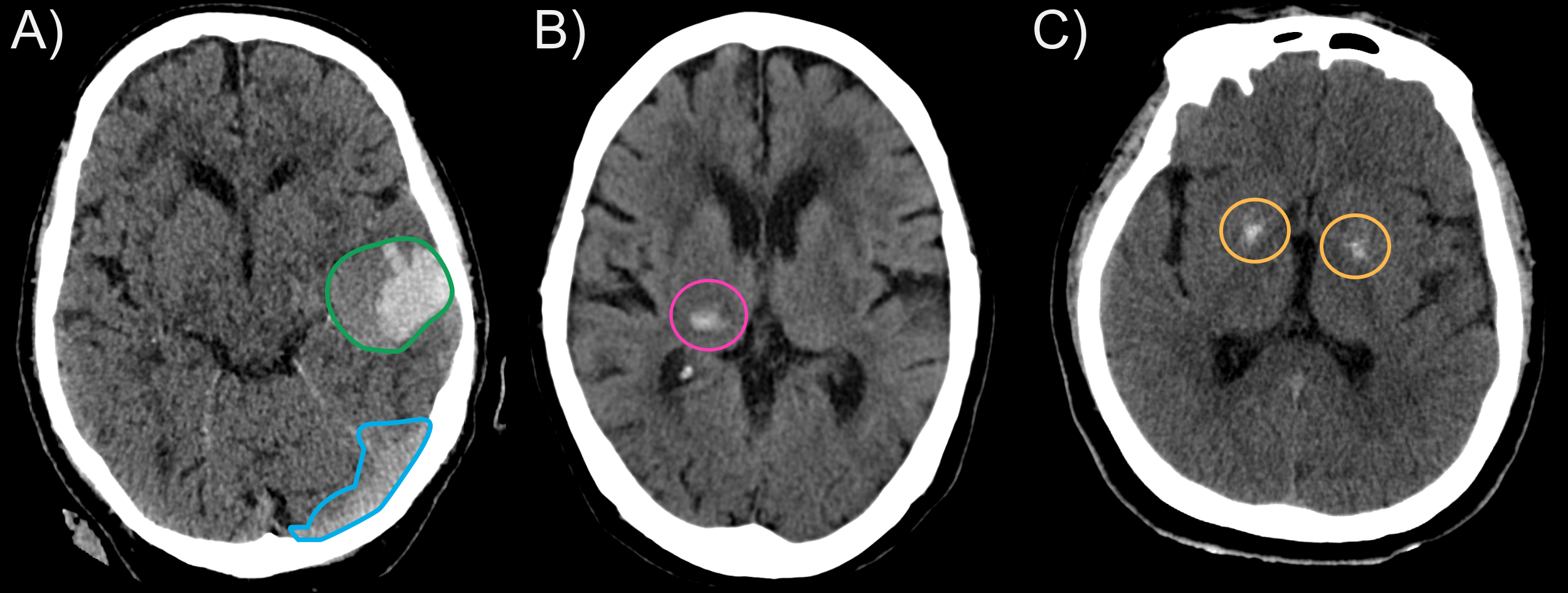}}
\end{minipage}

\caption{Insights into the challenges of brain hemorrhage detection with examples of hard cases. A) The patient suffers both from a \textcolor{ForestGreen}{bleeding} and a \textcolor{Cerulean}{thrombosis} further below. While the thrombosis has a similar appearance to the bleeding, the blood is still in a blood vessel and is not considered a bleeding. B and C) Both images show hyperdensities near the ventricle system. While the spot in image B is a \textcolor{magenta}{bleeding}, the two spots in C are simple \textcolor{Dandelion}{calcifications} of the brain tissue.
}
\label{fig:insights}

\end{figure}

Trauma remains one of the leading causes of death around the world, despite recent improvement in treatment protocols \cite{Corbacioglu2018-jo}. The latest protocol consists of doing a whole-body CT of multi-trauma patients.
However, this also means that the radiologist has more anatomies to review and less time can be allocated for specific body parts. 

Medical students are often left alone to do these readings due to staff shortages and will often miss lesions \cite{Cohn2010TheIS} or will struggle on hard cases (see Fig. \ref{fig:insights}). 
In particular, missed intracranial hemorrhages can have devastating effects on the patient's odds of survival and future physical autonomy \cite{Wang2008-hg}. 
As a result, there is a clear utility in having expert tools that can focus on a specific part of the whole-body CT and highlight relevant areas.

When it comes to stroke, the majority of previous work on 3D data has been dedicated to semantic segmentation \cite{Wagner2023ArtificialII}. Segmentation masks take a long time to annotate and only experts have the required skills. \hl{Further,} given the diversity of strokes in terms of lesion size, location, and shape \cite{Wardlaw2003-sx}, state-of-the-art models only achieve  0.6 to 0.7 Dice score, which is not yet clinically sufficient \cite{li2023stateoftheart}.

Additionally, Dice score as a metric is ill-suited for detection, \hl{as it only captures the proportion of blood volume detected. Similarly, a high Hausdorff distance could be due to a badly segmented bleeding or by a missed one. Both metrics provide no information on how many bleedings are completely missed. In contrast, bleeding detection aims to detect and localize individual bleeding to aid the clinician in finding relevant injuries or complication in a patient. This is particularly important for multi-trauma patients, as they will frequently suffer from multiple ailments and hemorrhages.} As there are multiple solutions for annotating 3D voxel data, we take a deep dive into the pros and cons of different methods. As this process is often tedious, one can be tempted to choose a fast but inaccurate solution. As such, we also analyze the effect of various annotation errors on the model performance.

Detecting bleeding of vastly different size, shape, and position requires a multiscale approach. Inspired by Feature Pyramid Network (FPN) strategy of Retina-Net \cite{10.1007/978-3-030-87240-3_51,lin2018focal}, we introduce an anisotropic-resolution-aware method leveraging features from 5 different, axis-independent scales.
Additionally, we introduce a family of anchors based on the distribution of the bleeding aspect ratio in training data, that also respect the anatomical structure of the brain. However, the network still needs to be able to learn these bleeding-specific shapes. To this end and inspired by the Complete Iou loss \cite{zheng2019distanceiou}, we propose a novel Voxel-Complete IoU (\textbf{VC-IoU}) loss that encourages the network to learn the \textbf{3D aspect ratios} of bounding boxes. This loss specifically penalizes deviations in aspect ratios along all three planes in the volume. We evaluate our proposed method on the publicly available INSTANCE2022 dataset \cite{li2023stateoftheart}, as well as on a private cohort for external validation.
 Our contributions in this work include:
\begin{enumerate}
    \item A novel loss (\textbf{VC-IoU}) for object detection in 3D voxel images\footnote{Code available at \url{https://github.com/MECLabTUDA/VoxelSceneGraph}}, which leads to more precise detections.
    \item An in depth comparison of annotation effort of multiple methods and an assessment of the impact of noisy bounding box annotation on the model's performance. These insights can help research groups to select the most adequate solution for other applications.

\end{enumerate}

%% file: Chapters/2_methods.tex
\section{Methodology}
\label{sec:meth}

In this section, we introduce our method for bleeding detection. More precisely, we introduce our backbone architecture, then define our novel Voxel-Complete IoU (VC-IoU) loss and finally go over challenges with sampling.

\subsection{Backbone architecture}

\begin{figure}

\begin{minipage}{1.0\linewidth}
  \centering
  \centerline{\includegraphics[width=10cm]{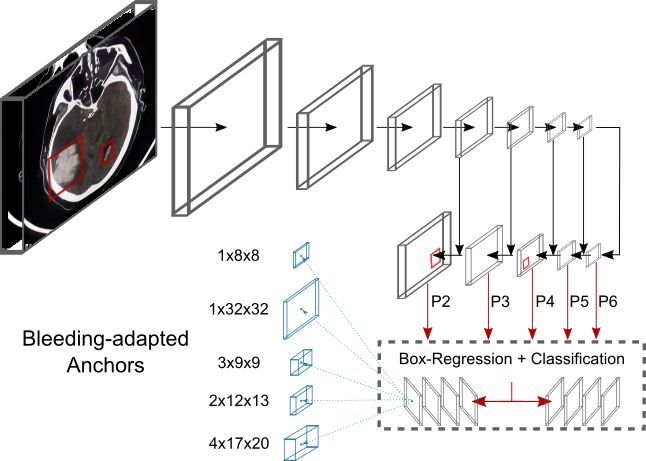}}
\end{minipage}

\caption{Our method for bleeding detection. Anchor sizes are given for level P2.}
\label{fig:fpn}

\end{figure}

The proposed 3D object detection method consists of a 3D Retina-Net with a Resnet-50-based FPN. This architecture has already proved its usefulness for 3D medical imaging \cite{DBLP:journals/corr/abs-1811-08661} as its FPN allows to leverage multiscale features. In particular, it is very flexible regarding the pyramid levels used. \textbf{We choose to not only use levels P2 to P5, but to also include P6 for detection} (see Fig. \ref{fig:fpn}). This is crucial as the volume of the bleeding can range from only 0.1cm$^3$ to more than 100cm$^3$. 
Additionally thanks to the convolutional nature of this architecture, we can customize the in-slice and depth-wise resolution scaling used to compute the next pyramid level. For instance, we use an in-slice downscaling factor of 2 from P0 to P1 and from P1 to P2. So if input slices have a 512\x512px size, their shape will already be shrunk to 128\x128px at P2. 
We only reduce the depth-wise resolution from P3 to P4 and from P4 to P5, since the slice thickness is 10\x\ greater than the in-slice resolution. 

One also has to consider the great diversity in bleeding shape when designing the shape of the anchors at a 5mm slice thickness. Indeed, small bleeding may appear in one up to four slices. This range in object depth is problematic for matching, as the Intersection over Union (IoU) of a potential match with a ground truth box can be excessively penalized by an inadequately chosen depth for the designed anchors. 
While the ATSS matching algorithm \cite{zhang2020bridging} does not require a hard IoU threshold for matching, the convergence rate of the box regression head can still suffer from misshapen anchors.
 \hl{Increasing the number of anchors can potentially improve matching results, but also increases the memory needed for computation. We have iteratively selected a family of 5 anchors. These are representative of the diverse 3D aspect ratios of bleeding within the datasets, while limiting the hardware requirements.}

\subsection{Voxel-Complete IoU loss}

\begin{figure}[t]

\begin{minipage}{1.0\linewidth}
  \centering
  \centerline{\includegraphics[width=10cm]{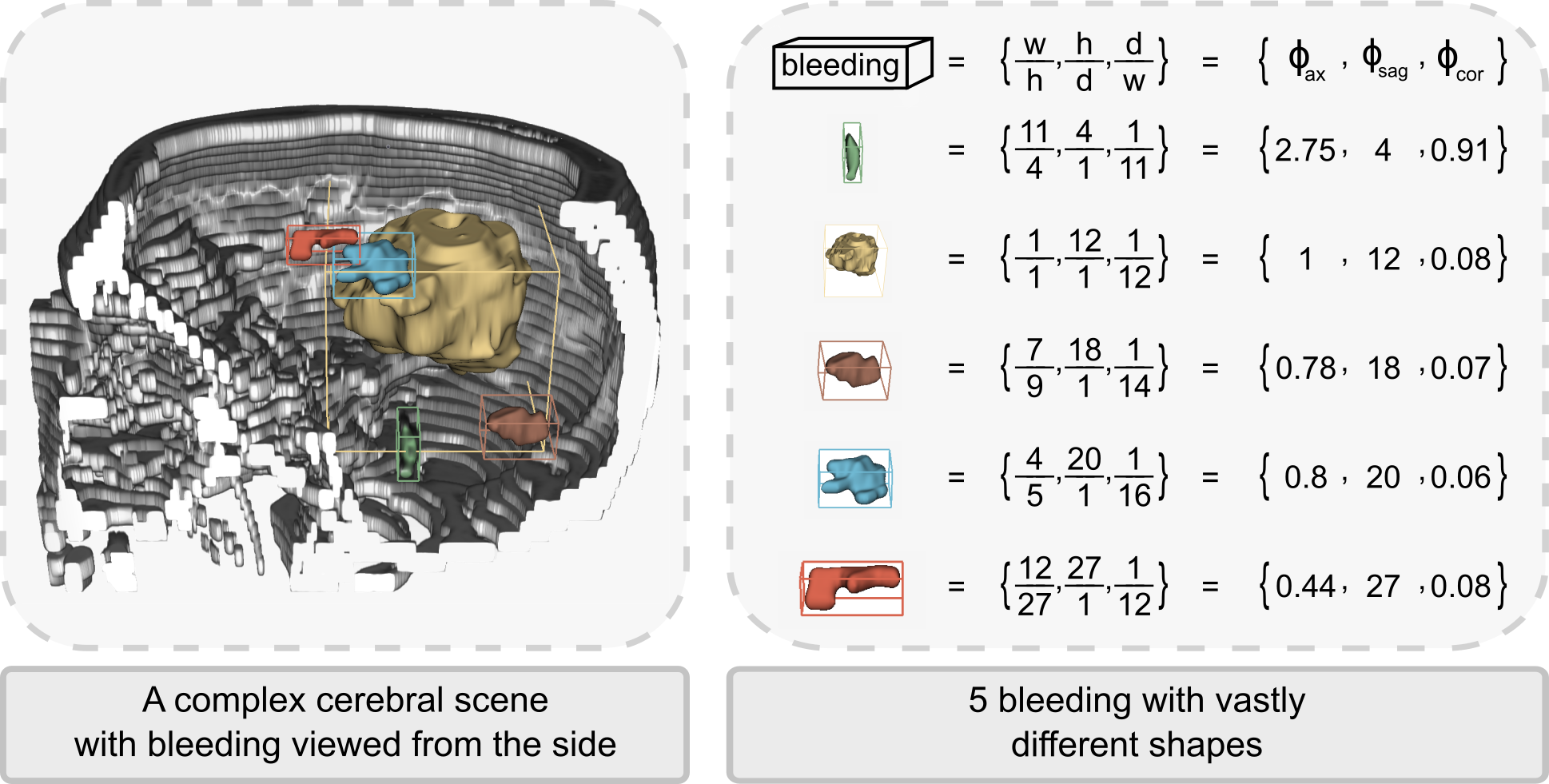}}
\end{minipage}

\caption{In 2D, the aspect ratio $\phi$ gives a scale-independent representation of a given bounding box. However, no such representation exists for 3D bounding boxes. We can instead consider the three aspect ratios $\{\phi_{ax}, \phi_{sag}, \phi_{cor}\}$ in the three natural planes of the 3D space (axial, sagittal and coronal). In particular, bleeding have a wide variety of shapes even \hl{within a single individual}.
}
\label{fig:aspect_ratios}

\end{figure}

Learning these specific shapes is not trivial. For 2D boxes, the aspect ratio is commonly defined as the proportion between the box's width and its height. However, no single ratio can be defined for a given \hl{3D} box (see Fig. \ref{fig:aspect_ratios}). Inspired by Zheng et al. \cite{zheng2019distanceiou}, we design a novel Voxel-Complete IoU (\textbf{VC-IoU}) loss by measuring the consistency of all three aspect ratios for a predicted box $B=(w,h,d)$ and a ground truth box $B^{gt}=(w^{gt},h^{gt},d^{gt})$:

\begin{equation*}
    v = \frac{4}{\pi^2}
       \left[
    \begin{aligned}
    & (\arctan\frac{w^{gt}}{h^{gt}} - \arctan\frac{w}{h})^2\\
    + & (\arctan\frac{h^{gt}}{d^{gt}} - \arctan\frac{h}{d})^2\\
    + & (\arctan\frac{d^{gt}}{w^{gt}} - \arctan\frac{d}{w})^2\\
        \end{aligned}
    \right]
\end{equation*}

Similarly to the C-IoU loss \cite{zheng2019distanceiou}, we define a trade-off factor $\alpha$ as

\begin{equation*}
    \alpha = \frac{v}{1 - IoU + v}
\end{equation*}

to ensure that the overlap factor is given priority. The VC-IoU loss can then be defined as
\begin{equation*}
    \mathcal{L}_{VC-IoU} = \mathcal{L}_{DIoU} + \alpha v
\end{equation*}
where $\mathcal{L}_{DIoU}$ is the Distance IoU \cite{zheng2019distanceiou}.  
There are of course multiple options to aggregate the regularization term for all three aspect ratios when defining $v$. We choose to compute a sum rather than a mean, as we consider each of these penalties to be as important as the single penalty that would be observed in a 2D setting. Besides, the acceptable IoU threshold is typically much lower in 3D applications than in 2D ones (e.g. 10\% vs 50\%). So, having higher values of $v$ allows the trade-off factor $\alpha$ to be in a comparable range compared to a 2D CIoU loss.

\subsection{Sampling of bleeding}

The sampling algorithm for the classification head has to be tuned to avoid detecting all hyperdensities within the head. In particular, the transverse sigmoid sinuses are large veins along the back of the skull that has a texture close to bleeding in CT imaging. It is critical that a model learns to distinguish such structures. As such, it is critical to select negative samples from the most confidently predicted false positives. Qualitative examples are shown in Fig. \ref{fig:insights}.

%% file: Chapters/3_data.tex
\newpage
\section{Data}
\label{sec:data}

While whole-body CT are acquired for trauma patients, the focus for this application is exclusively on the patient's head. 
Handling the head region of whole-body CT or directly using a head CT are equivalent, as converting the former to the latter only requires a rough localization of the patient's neck. 
Since it is not a hard task, we consider it out of scope of this study and directly use head CTs for this study.

\subsection{Source images}

\begin{figure}

\begin{minipage}{1.0\linewidth}
  \centering
  \centerline{\includegraphics[width=\linewidth]{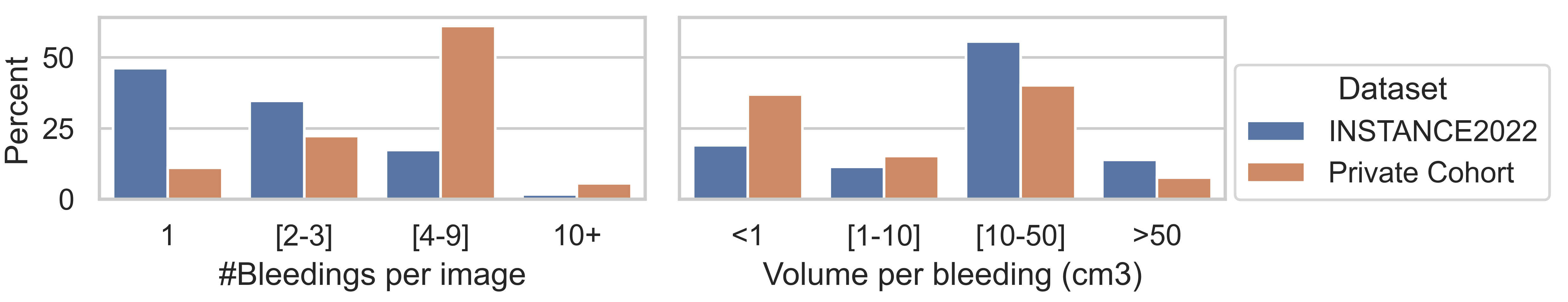}}
\end{minipage}

\caption{\hl{Distribution shift of the number of bleeding per image (left) and volume per bleeding (right) between the INSTANCE2022 dataset and our private cohort.}
}
\label{fig:dataset}

\end{figure}

\begin{figure}

\begin{minipage}{1.0\linewidth}
  \centering
  \centerline{\includegraphics[width=11cm]{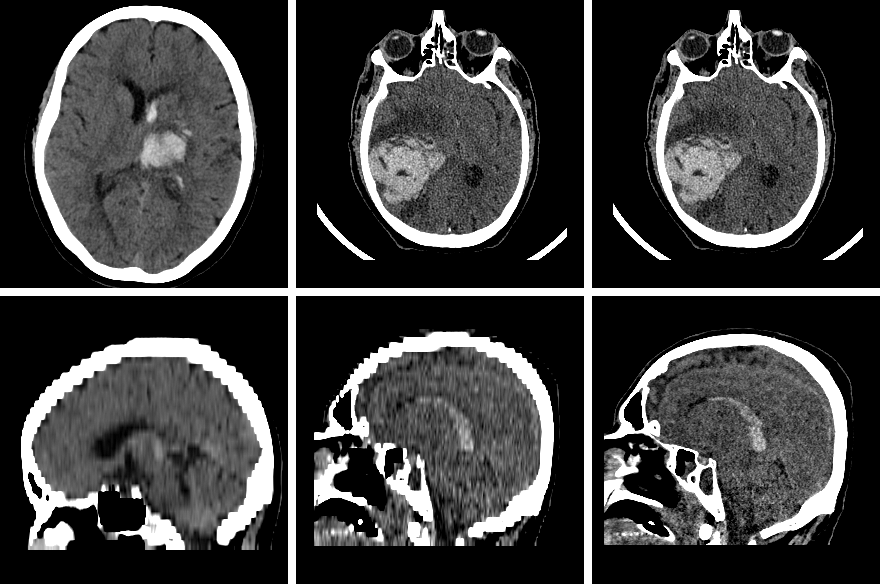}}

\end{minipage}

\caption{Visualization of head CTs, axial (top) and sagittal (bottom). INSTANCE2022 data (left). Downscaling (middle) of the  original private cohort data (right).}
\label{fig:res_comp}

\end{figure}

The first dataset is the publicly available INSTANCE2022 challenge dataset \cite{li2023stateoftheart}. 
It contains 130 non-contrast head CTs with 0.42\x0.42\x 5mm$^3$ voxel spacing from patients diagnosed with intracranial hemorrhage. 
Additionally, we use a private dataset for the purpose of external validation.
It is constituted of 18 head CTs of patients diagnosed with intracranial hemorrhage between 2021 and 2022. As shown in \hl{Fig.} \ref{fig:dataset}, this dataset contains out-of-distribution cases, both in terms of bleeding size and number.
These images are high-resolution head CTs and have a native voxel spacing of 0.41\x0.41\x0.3mm$^3$.
As such, we resampled them  to a voxel spacing of 0.41\x0.41\x5mm$^3$ for inference (see Fig.\ \ref{fig:res_comp}), to match the resolution of the training data.\\

\subsection{Data annotation}

Of the 130 images from INSTANCE2022, 100 were released as training cases along with their corresponding segmentation of the bleeding. 
We derived the object annotation by generating the 3D connected components from the masks and computed the bounding box of each component. All other images were annotated internally by a senior radiologist from the University Medical Center Mainz using 3D Slicer \cite{Fedorov2012-qh}. In particular, we chose to annotate the objects as rough label maps (see section \ref{sec:annot_methods}).

%% file: Chapters/4_exp.tex
\section{Experimental Setup}
\label{sec:exp}
In this section, we introduce the base images used and their annotation process.
We then describe our evaluation setup for the different stages of our method.

\subsection{Bounding box prediction}
We compare our method to the state-of-the-art nnDetection framework \cite{10.1007/978-3-030-87240-3_51}.
To train the detection models, we split the first 100 cases from the INSTANCE2022 dataset \cite{li2023stateoftheart} for training and validation in an 80/20 fashion. The 30 remaining cases are used for in-distribution testing. Finally, the private cohort is used for external validation.

\subsection{Segmentation for object detection}

To evaluate segmentation for bleeding detection, we use a nnU-Net \cite{Isensee2021} trained with 5-fold cross-validation. This architecture performed well in the test phase of INSTANCE2022 with a Dice score of 0.69 \cite{li2023stateoftheart}. 
We then utilize the sister framework nnDetection \cite{10.1007/978-3-030-87240-3_51} to ensemble and convert the predicted segmentations to bounding boxes. The in-distribution testing and external validation are done in the same manner as for bounding box prediction.

\subsection{Metrics}

We evaluate methods for bleeding detection using average precision (AP) and average recall (AR) at IoU thresholds of 10\% and 30\%. Prior studies \cite{10.1007/978-3-030-87240-3_51} suggest that using a threshold of 10\% IoU is sufficient, as IoU is more penalizing in 3D and clinical applications only require a coarse localization. However, having such a low threshold is problematic for bleeding detection, as their volume can vary greatly. In particular, a bounding box predicted for a given bleeding could easily overlap with a smaller neighboring bleeding. At a low 10\% IoU threshold, this bounding box can often sufficiently overlap the second bleeding to be counted as a positive match for both objects. This can be particularly troublesome if the first larger bleeding was in fact a false positive (see Fig. \ref{fig:insights} A)). To mitigate this issue, we also provide all results at a 30\% IoU threshold.

\subsection{Implementation details}

Our framework is implemented in Python 3.10 and PyTorch 1.13 \cite{paszke2019pytorch}. \hl{All configuration files including hyperparameters and data splits will be made available with the code. All models are trained on a single  RTX3090 GPU. The training for our method takes approximately 7 hours and requires up to 16 GB of VRAM. As a comparison, \textit{nnDetection}} \cite{10.1007/978-3-030-87240-3_51} \hl{takes 24 hours to train over 5 folds and requires 24 GB of VRAM. This increased memory usage compared to our method can mainly be attributed to the higher number of anchors that \textit{nnDetection} uses.}

%% file: Chapters/5_results.tex
\section{Results}
\label{sec:res}

In this section, we first compare our method to the existing state-of-the-art methods. We then further evaluate our method through ablation studies. Finally, since annotating data is performed under heavy resource constraints in the clinical world, we perform a comparison of different annotation solutions and study the impact of imprecise boxes on model performances.

\subsection{Bleeding detection}
\label{sec:detec}

\begin{table}
\centering
\begin{tabular}{ l  c  c  c  c  c  c  c  c}
\toprule
& \multicolumn{4}{c}{INSTANCE2022} & \multicolumn{4}{c}{Private Cohort}\\
\cmidrule(lr){2-5}\cmidrule(lr){6-9}
Method &  AR$_{10}\uparrow$ & AR$_{30}\uparrow$ & AP$_{10}\uparrow$ & AP$_{30}\uparrow$ & AR$_{10}\uparrow$ & AR$_{30}\uparrow$ & AP$_{10}\uparrow$ & AP$_{30}\uparrow$\\
\midrule
nnU-Net$^{1}$ & 0.789 & 0.600 & 0.708 & 0.522 & 0.561 & 0.378 & 0.416 & 0.257\\
nnDetection & 0.815 & 0.631 & 0.672 & 0.549 & 0.704 & 0.439 & 0.494 & 0.263\\
Ours & \bf 0.892 & \bf 0.877 & \bf 0.760 & \bf 0.728 & \bf 0.724 & \bf 0.653 & \bf 0.590 & \bf 0.514\\
\bottomrule
\end{tabular}

\caption{Detection rates at 10\% and 30\% IoU using our method, nnDetection \cite{10.1007/978-3-030-87240-3_51} and $^{1}$nnU-Net for detection \cite{10.1007/978-3-030-87240-3_51}.
}
\label{detec}
\end{table}

\begin{figure}
\begin{minipage}{1.0\linewidth}
  \centering
  \centerline{\includegraphics[width=\linewidth]{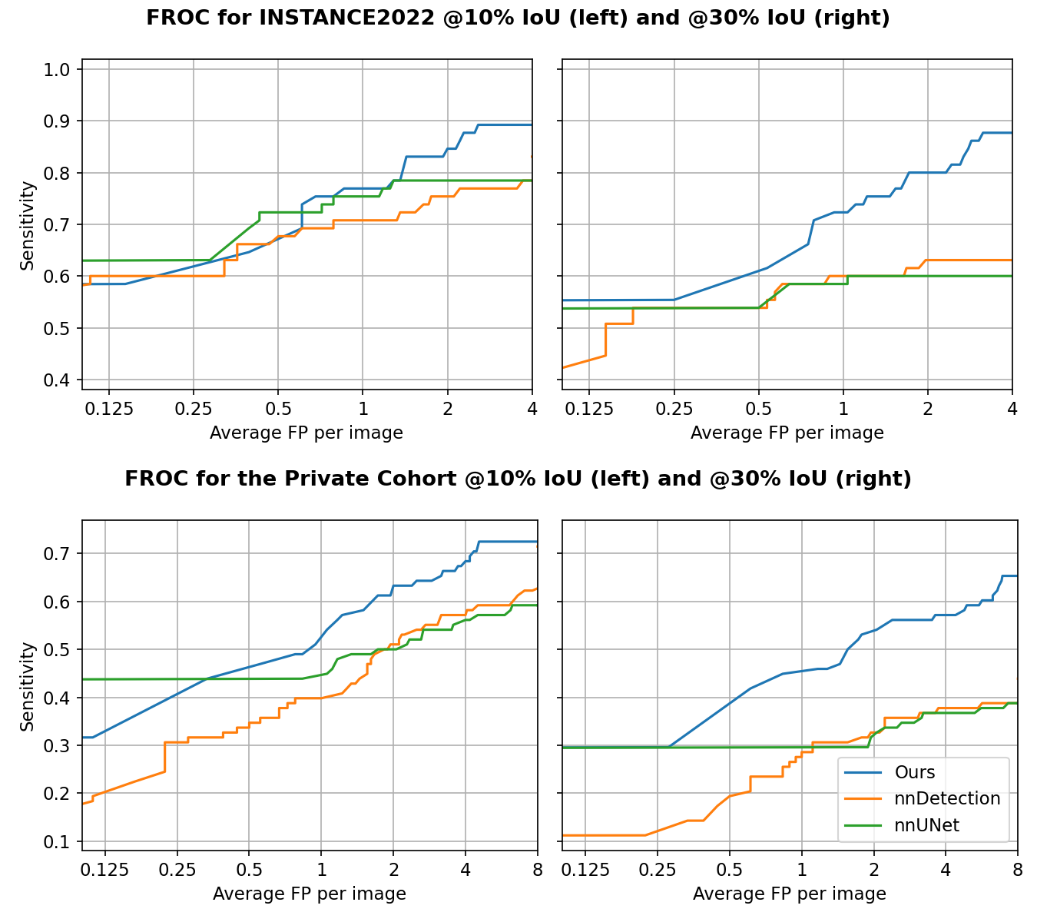}}
    \caption{FROC analysis of our method, nnDetection \cite{10.1007/978-3-030-87240-3_51} and $^{1}$nnU-Net for detection \cite{10.1007/978-3-030-87240-3_51} on the INSTANCE2022 dataset (top), and our private cohort (bottom) at 10\% IoU (left) and 30\% IoU (right).}
    \label{fig:FROC}
\end{minipage}
\end{figure}

\begin{figure}[t]
\begin{minipage}{1.0\linewidth}
  \centering
  \centerline{\includegraphics[width=11cm]{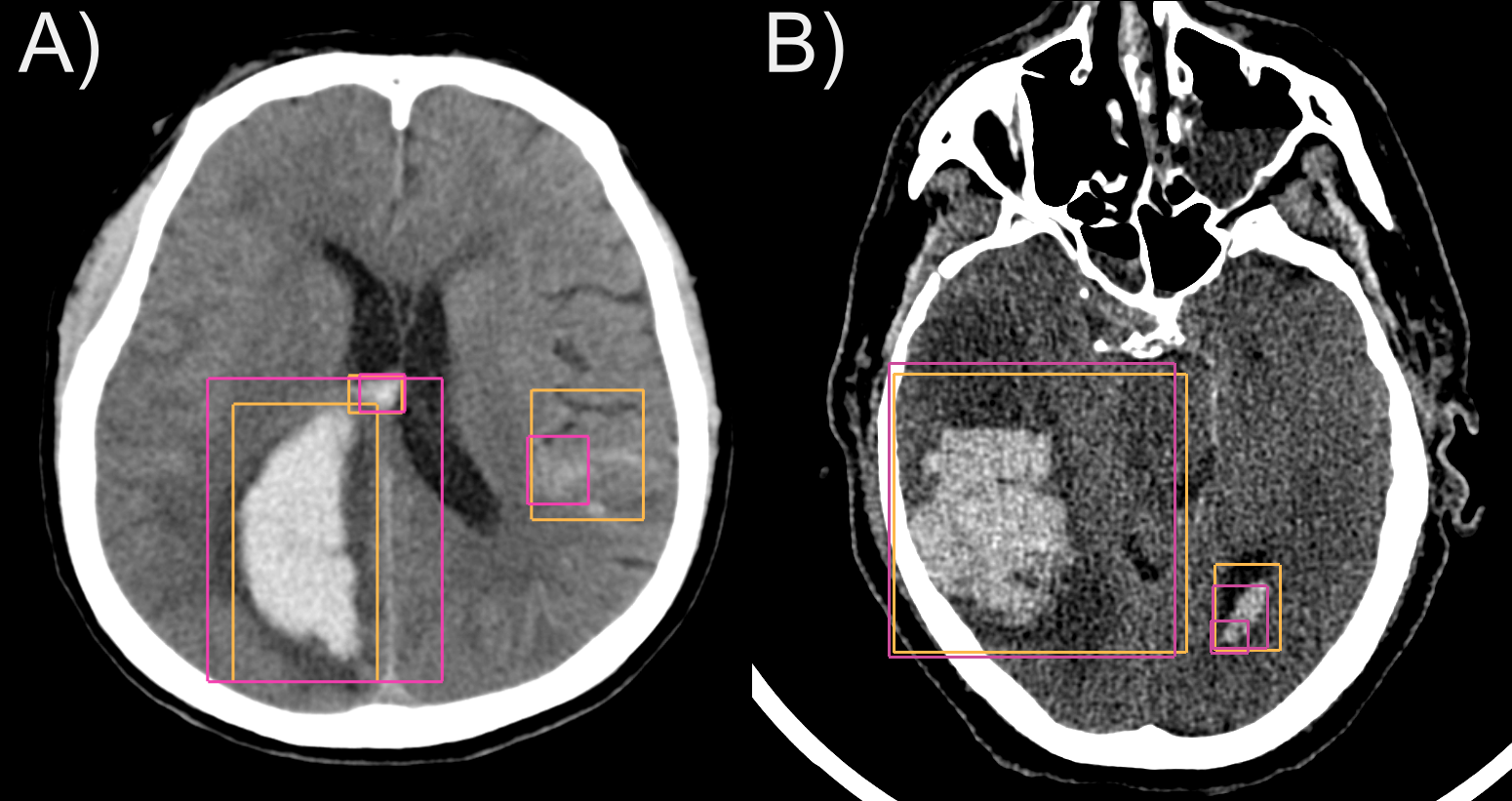}}

    \caption{
    Detection examples using \textcolor{magenta}{our method} on A) INSTANCE2022 and B) the private cohort. \textcolor{Dandelion}{Ground truth} boxes are also provided for comparison. In A), the model correctly identifies the three bleeding present. In particular, the smaller center bleeding is in the patient's ventricle system, which can cause serious complications if left untreated \cite{Garton2016-zg}. The model also manages to detect the diffuse bleeding to the right. In B), the model again detects relevant bleeding. Please note that \textcolor{Dandelion}{ground truth} boxes appear larger than the bleeding within this slice, as these expand further in neighboring slices. The bleeding to the bottom right also highlights the challenges of non-maxima suppression, as it gets detected twice but with little overlap between \textcolor{magenta}{predictions}.
    }

    \label{fig:qual_res}
\end{minipage}
\end{figure}

We first evaluate our method against state-of-the-art solutions. As shown in Table \ref{detec}, \textbf{our method significantly outperforms both nnDetection and nnU-Net on both datasets}. nnU-Net comes last, showing how segmentation is inadequate for object detection. While there is a drop in performance on the private cohort for all methods, ours is \hl{more robust against distribution shifts}. The FROC analysis (Fig. \ref{fig:FROC}) sheds some further light on the difference in performance between datasets for each method. 
Additionally, we provide some qualitative results for our method in Fig. \ref{fig:qual_res}.

\newpage
\subsection{Ablation: loss function}

To detail the influence of the additional loss term in our VC-IoU loss, we plug commonly-used losses in our model architecture and compare the results. Table \ref{losses} confirms that our VC-IoU allows the network to detect more bleeding more precisely, with a relative increase of 5\% Average Recall for both IoU thresholds and for both datasets. 

\begin{table}
\centering
\begin{tabular}{l  c  c  c  c  c  c  c  c}
\toprule
& \multicolumn{4}{c}{INSTANCE2022} & \multicolumn{4}{c}{Private Cohort}\\
\cmidrule(lr){2-5}\cmidrule(lr){6-9}
Loss &  AR$_{10}\uparrow$ & AR$_{30}\uparrow$ & AP$_{10}\uparrow$ & AP$_{30}\uparrow$ & AR$_{10}\uparrow$ & AR$_{30}\uparrow$ & AP$_{10}\uparrow$ & AP$_{30}\uparrow$\\
\midrule
Smooth L1 \cite{girshick2015fast} & 0.846 & 0.754 & 0.660 & 0.532 & \bf 0.724 & 0.531 & 0.460 & 0.311\\
DIoU \cite{zheng2019distanceiou} & 0.877 & 0.815 & \bf  0.758 & 0.692 & 0.684 & 0.622 & \bf 0.592 & \bf  0.515\\
VC-IoU (Ours) & \bf 0.892 & \bf 0.877 & \bf 0.760 & \bf 0.728 & \bf 0.724 & \bf 0.653 & \bf 0.590 & \bf 0.514\\
\bottomrule
\end{tabular}

\caption{\hl{Ablation: Detection rates at 30\% IoU using our method and different loss functions for bounding box regression}.
}
\label{losses}
\end{table}

\subsection{Ablation: bleeding size}

This ablation aims to shed light on the drop of performance of our method on the private cohort. In particular, one might recall that the bleeding size distribution is significantly different from the distribution in the training set. As such, we additionally give the performance of our method on each size group (Table \ref{size}). We can observe the steepest drop in performance from INSTANCE2022 to the private cohort for both the smallest and largest bleeding. The former can be expected, as smaller bleeding are harder to detect and are more prevalent in the private cohort. However, the latter phenomenon is due to a strong distribution shift of the cerebral scene. Indeed, the private cohort also contains trauma patients with severe hemorrhages but also fractures or even an open skull. No similar cases are present in the training set.

\begin{table}
\centering
\begin{tabular}{ l  c  c  c  c  c  c  c  c }
\toprule
& \multicolumn{4}{c}{INSTANCE2022} & \multicolumn{4}{c}{Private Cohort}\\
\cmidrule(lr){2-5}\cmidrule(lr){6-9}
Bleeding size (cm$^3$) &  AR$_{10}\uparrow$ & AR$_{30}\uparrow$ & AP$_{10}\uparrow$ & AP$_{30}\uparrow$ & AR$_{10}\uparrow$ & AR$_{30}\uparrow$ & AP$_{10}\uparrow$ & AP$_{30}\uparrow$\\
\midrule
all & 0.892 & 0.877 & 0.760 & 0.728 & 0.724 & 0.653 & 0.590 & 0.514\\
$<1$ & 0.824 & 0.765 & 0.605 & 0.585 & 0.611 & 0.528 & 0.521 & 0.481\\
$[1\text{-}10]$ & 0.895 & 0.895 & 0.857 & 0.836 & 0.788 & 0.727 & 0.713 & 0.637\\
$[10\text{-}50]$ & 0.867 & 0.867 & 0.784 & 0.771 & 0.571 & 0.571 & 0.475 & 0.466\\
$>50$ & 0.929 & 0.929 & 0.896 & 0.865 & 0.588 & 0.471 & 0.497 & 0.373\\
\bottomrule
\end{tabular}

\caption{Detection rates at 10\% and 30\% IoU using our method for each bleeding size group.
}
\label{size}

\end{table}

\subsection{Comparison of annotation methods}
\label{sec:annot_methods}

Annotating 3D medical images is a time intensive task. Tools like 3D Slicer \cite{Fedorov2012-qh} offer multiple solutions, both for bounding box annotation and segmentation. We consider the 4 following methods:

\begin{enumerate}
    \item Directly placing bounding boxes within 3D Slicer.
    \item Segmenting a rough and sparse label map, from which bounding boxes are computed for each segment. This method does not require the complete segmentation of a bleeding, only a bleeding's extremities have to be segmented for the resulting bounding box to be accurate. 
    \item Precisely segmenting the bleeding.
    \item Using a Deep Learning model to pre-segment the images and manually correcting the resulting masks.
\end{enumerate}

\begin{figure}
\begin{minipage}{1.0\linewidth}
  \centering
  \centerline{\includegraphics[width=11cm]{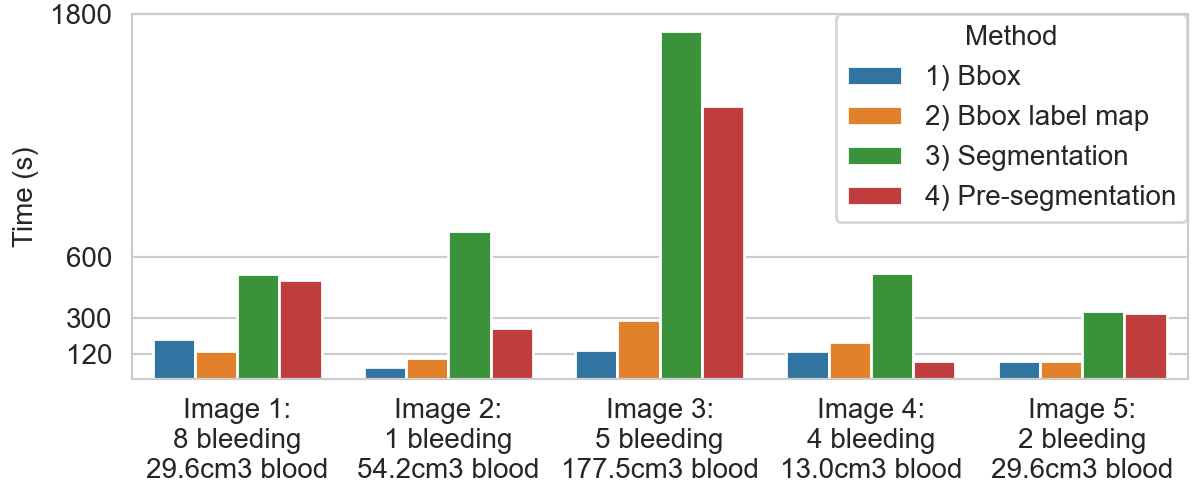}}
    \caption{Annotation effort in seconds to annotate 5 downsampled images from our private cohort using 3D Slicer with 4 different methods. 1) Annotating bounding boxes directly 2) Annotating bounding boxes through rough label maps 3) Segmenting from scratch 4) Refining of automatically-generated pre-segmentation.}
    \label{fig:times}
\end{minipage}
\end{figure}

We evaluate these 4 methods on 5 images of our private cohort with 5mm slice-thickness and use the model described in section \ref{sec:exp} for the pre-segmentation. It is critical to choose a single method before starting the annotation process, as discrepancies can arise, resulting in lower detection rates \cite{9857206}.
As shown in Fig. \ref{fig:times}, bounding boxes are 2 to 6 times faster to annotate than segmentation masks. Annotating a label map is slightly slower but more precise, as object bounds are still accurately annotated. Additionally, it provides a clearer overview of which objects have been annotated.

In contrast, segmenting images is always much slower, even for smaller bleeding. However, it is now unrealistic to expect the expert to annotate images from scratch, when Deep Learning models can pre-segment to an extent. The most tedious part of segmenting is making sure that the object boundaries are accurate. As such, the speed-up offered by the pre-segmentation is not determined by the volume of the pre-segmented mask. The pre-segmentation of image 3 is a good example of this. Even though 70\% of the bleeding volume was already pre-segmented, many boundaries still had to be adjusted.

\subsection{Impact of noisy annotation}
\label{sec:noise}

Annotating images is a task often done after work hours. As such, it is unsurprising when some noise is introduced in the annotated bounding boxes. We evaluate its impact on the detection rate by simulating the following scenarios:

\begin{enumerate}
    \item Bounding boxes are too \textcolor{ForestGreen}{small}. We randomly shrink all bounding boxes by up to 10\%. The resulting boxes have on average 78.5\% IoU with their original counterparts.
    \item Bounding boxes are too \textcolor{Cerulean}{large}. We randomly enlarge all bounding boxes by up to 10\%. The resulting boxes have on average 82.2\% IoU with their original counterparts.
    \item Bounding boxes are \textcolor{magenta}{off center}. We randomly move all bounding boxes' center by up to 10\% of their size. The resulting boxes have on average 79.9\% IoU with their original counterparts.
    \item Some bleeding are \textcolor{Dandelion}{not annotated}, especially smaller ones. We randomly remove 20\% of bleeding under 1 cm$^3$ and 10\% of bleeding under 10 cm$^3$.
\end{enumerate}

\begin{table}
\centering
\begin{tabular}{ l  c  c  c  c  c  c  c  c}
\toprule
& \multicolumn{4}{c}{INSTANCE2022} & \multicolumn{4}{c}{Private Cohort}\\
\cmidrule(lr){2-5}\cmidrule(lr){6-9}
Noise &  AR$_{10}\uparrow$ & AR$_{30}\uparrow$ & AP$_{10}\uparrow$ & AP$_{30}\uparrow$ & AR$_{10}\uparrow$ & AR$_{30}\uparrow$ & AP$_{10}\uparrow$ & AP$_{30}\uparrow$\\
\midrule
None & 0.892 & 0.877 & 0.760 & 0.728 & 0.724 & 0.653 & 0.590 & 0.514\\
\textcolor{ForestGreen}{Smaller} & 0.862 & 0.815 & 0.763 & 0.706 & 0.735 & 0.643 & 0.586 & 0.463\\
\textcolor{Cerulean}{Larger} & 0.862 & 0.785 & 0.767 & 0.682 & 0.704 & 0.602 & 0.584 & 0.464\\
\textcolor{magenta}{Moved} & 0.862 & 0.831 & 0.758 & 0.710 & 0.724 & 0.622 & 0.605 & 0.500\\
\textcolor{Dandelion}{Missing} & 0.862 & 0.831 & 0.766 & 0.732 & 0.643 & 0.571 & 0.528 & 0.453\\
\bottomrule
\end{tabular}

\caption{Detection rates at 10\% and 30\% IoU using our method under different annotation noise regimes in the training data (bottom, see section \ref{sec:noise}). 
}
\label{noise}
\end{table}

The results in Table \ref{noise} show that even slight annotation errors in bounding box size will hinder the network from the learning precise detections. Additionally, missing bleeding during the annotation process is most detrimental to the robustness of the network. Nevertheless, these models still outperform both nnDetection and nnU-Net \cite{10.1007/978-3-030-87240-3_51}.

In particular, if we consider which annotation method to use, one can better understand the trade-off of annotating bounding boxes directly. While images can be swiftly annotated, even slight errors can be detrimental to the final model performance. In contrast, annotating rough label maps solves these risks with only little effort overhead.

%% file: Chapters/6_discussion.tex
\section{Conclusion}

Detecting bleeding of vastly different size, shape, and position in 3D voxel imaging requires a multiscale approach. 
We introduce an anisotropic-resolution-aware method and a family of anchors that respect the anatomical structure of the brain. Additionally, we propose a novel Voxel-Complete IoU (VC-IoU) loss that encourages the network to learn the 3D aspect ratios of bounding boxes, which we evaluate our method on two datasets for brain hemorrhage detection. We demonstrate that our model significantly outperforms state-of-the-art methods \cite{10.1007/978-3-030-87240-3_51} and that our loss yields a relative increase in Average Recall of 5\% compared to other loss functions. \hl{Our method has the potential to provide a second level of security to the neuro-radiologist when reading the CT scan of a newly admitted multi-trauma patient and to help ensure that no bleeding is missed.}

As little data is currently publicly available for 3D object detection, training this method for any new applications would require to annotate new data. As annotation resources are limited in the clinical setting, we evaluate the cost of different annotation methods, as well as the impact of imprecise bounding boxes in the training data on the detection performance. These results can help shed light on whether segmentation or pure object detection is the better approach for new applications. With this work, we pave the way towards automated tools to offer strong decision support.

%% file: Chapters/7_ethics.tex
\section{Compliance with ethical standards}
\label{sec:ethics}

This study was performed in line with the principles of the Declaration of Helsinki. The retrospective evaluation of imaging data from the University Medical Center Mainz was approved by the local ethics boards (Project 2021-15948-retrospektiv). Ethical approval was not required, as confirmed by the license attached with the open access data.